\documentclass[conference]{IEEEtran}
\IEEEoverridecommandlockouts
\usepackage{cite}
\usepackage{amsmath,amssymb,amsfonts}
\usepackage{algorithmic}
\usepackage{subcaption}
\usepackage{graphicx}
\usepackage{hyperref}
\usepackage{multirow}  
\usepackage{float}
\usepackage{textcomp}

\usepackage{xcolor}
\def\BibTeX{{\rm B\kern-.05em{\sc i\kern-.025em b}\kern-.08em
    T\kern-.1667em\lower.7ex\hbox{E}\kern-.125emX}}
\begin{document}

\title{A Synthetic Data Pipeline for Supporting Manufacturing SMEs in Visual Assembly Control\\

}

\DeclareRobustCommand{\IEEEauthorrefmark}[1]{\smash{\textsuperscript{\footnotesize #1}}}

\author{\IEEEauthorblockN{
Jonas Werheid*\IEEEauthorrefmark{1},
Shengjie He\IEEEauthorrefmark{1}, 
Aymen Gannouni\IEEEauthorrefmark{1},
Anas Abdelrazeq\IEEEauthorrefmark{1},
Robert H. Schmitt\IEEEauthorrefmark{1}
}
\IEEEauthorblockA{\IEEEauthorrefmark{1}\textit{Chair of Intelligence in Quality Sensing, RWTH Aachen University, Aachen, Germany}}
\IEEEauthorblockA{\textit{* Correspondance: 
jonas.werheid@wzl-iqs.rwth-aachen.de}}
}

\maketitle

\begin{abstract}
Quality control of assembly processes is essential in manufacturing to ensure not only the quality of individual components but also their proper integration into the final product. To assist in this matter, automated assembly control using computer vision methods has been widely implemented. However, the costs associated with image acquisition, annotation, and training of computer vision algorithms pose challenges for integration, especially for small- and medium-sized enterprises (SMEs), which often lack the resources for extensive training, data collection, and manual image annotation. Synthetic data offers the potential to reduce manual data collection and labeling. Nevertheless, its practical application in the context of assembly quality remains limited. In this work, we present a novel approach for easily integrable and data-efficient visual assembly control. Our approach leverages simulated scene generation based on computer-aided design (CAD) data and object detection algorithms. The results demonstrate a time-saving pipeline for generating image data in manufacturing environments, achieving a mean Average Precision (mAP@0.5:0.95) up to 99,5\% for correctly identifying instances of synthetic planetary gear system components within our simulated training data, and up to 93\% when transferred to real-world camera-captured testing data. This research highlights the effectiveness of synthetic data generation within an adaptable pipeline and underscores its potential to support SMEs in implementing ressource-efficient visual assembly control solutions. 
\end{abstract}


\section{Introduction}
\label{Introduction}

An assembly process combines parts and materials involving machines, equipment, or workers in a defined sequence to create a finished product \cite{nof1997industrial}. Assembly control procedures play an important role in manufacturing environments to ensure the final product quality for customers \cite{Beiter00}. As human operators are labor-intensive and manual inspection is time-consuming when handling complex and repetitive tasks \cite{BBD22}, technology is often used to automate quality control procedures. For example, computer vision methods are common for visual quality control of parts or materials, and also whole assembly procedures \cite{ZAMORAHERNANDEZ2021103485}. To implement these methods effectively, a training process is required to enable the algorithms to recognize relevant patterns and detect defects. This training typically relies on large amounts of labeled image data, which are acquired through camera-based setups and annotated manually \cite{Basamakis22}.

However, camera-based image acquisition is resource-intensive, particularly due to the effort required for the manual annotation, and comes with additional limitations, such as inconsistent image quality and limited data diversity, especially in cases of imbalanced defect classes \cite{eng2.12910}. Therefore, for example, small- and medium-sized enterprises (SMEs) with limited technical and financial resources are often at a disadvantage when it comes to implementing computer vision technologies \cite{Werheid2025}.

In contrast to traditional data collection methods, recent state-of-the-art research explores synthetic data generation for a wide range of industrial tasks involving both discrete (e.g., binary data, point clouds) and continuous manufacturing process data types (e.g., time-series, images, video) \cite{Buggineni22}. For image data generation, various techniques are presented in research, including generative adversarial networks (GANs) \cite{Qian22}, domain randomization \cite{Ameperosa20}, and computer graphics tools, along with CAD files, as input data \cite{LAI202069}. Common computer graphics tools, such as Blender \cite{blender18}, can be used for synthetic data generation with CAD files for defect detection \cite{MONNET2024767, SONG2024109852}, object segmentation \cite{Károly22}, and pose tracking \cite{LV2024102788}. Within Blender, the open source library Blenderproc is frequently used in research to enable automated scene generation within Python code \cite{Den2023}.

In addition, previous work demonstrated the usage of Blender for synthetic data generation, particularly in the context of assembly tasks. For example, Conrad et al \cite{CONRAD2024239} used Blender for synthetic data generation to identify error states in manual cable assembly. In comparison to other solutions, various error states can be detected by the presented object detection approach \cite{CONRAD2024239}. Due to differences between simulations and the real world, a performance drop, commonly referred to as the Sim2Real gap, occurs when models trained on synthetic data are applied to real-world scenarios \cite{Biruduganti25}. \cite{Bai24} presented a quantitative evaluation of object detection and semantic segmentation tasks, reporting Sim2Real gaps for various synthetically created objects, with real-world performance differences ranging from 1.7\% to 37.9\%. Ratal et al \cite{RAWAL20251668} showed BlenderProc with domain adaption techniques to reduce the gap from synthetic training to real inference in visual assembly control. Wu et al \cite{WU2022138} proposed Blender-based approaches adapted to certain domains, such as disassembly of electric motors to enable easier direct transfer from simulation to reality. Another approach involved generating assembly instructions based on CAD data rendered in Blender \cite{BURGGRAEF2024775}. Existing approaches do not provide a complete, easily integrable assembly pipeline suitable for real-world adoption by SMEs. Existing approaches often lack an integrated and easily deployable pipeline for visual assembly control that is practical for smaller companies and their internal teams.

We present CAD-driven synthetic data generation with Blender and BlenderProc, linked to an open-source object detection model for automated data generation, annotation, and algorithm training as a synthetic data-based visual assembly control pipeline. Within an experimental use case, the pipeline is demonstrated for assembly control of planetary gear components, including individual component detection and verification of correct assembly combinations. Furthermore, to enable low-barrier adoption by SMEs and their internal teams, all components are seamlessly integrated into a single, ready-to-use pipeline, accompanied by source code, and video documentation.

Firstly, Section \ref{pipeline} describes the synthetic data-based pipeline, and Section \ref{setup} presents its application within an experimental use case. Subsequently, Section \ref{results} presents the results alongside a discussion. Finally, Section \ref{conclusion} concludes the work, outlines its impact for SMEs, and discusses future directions.

\section{Synthetic Data-Based Pipeline}
\label{pipeline}
 
\begin{figure}
    \centering
        \includegraphics[width=\linewidth]{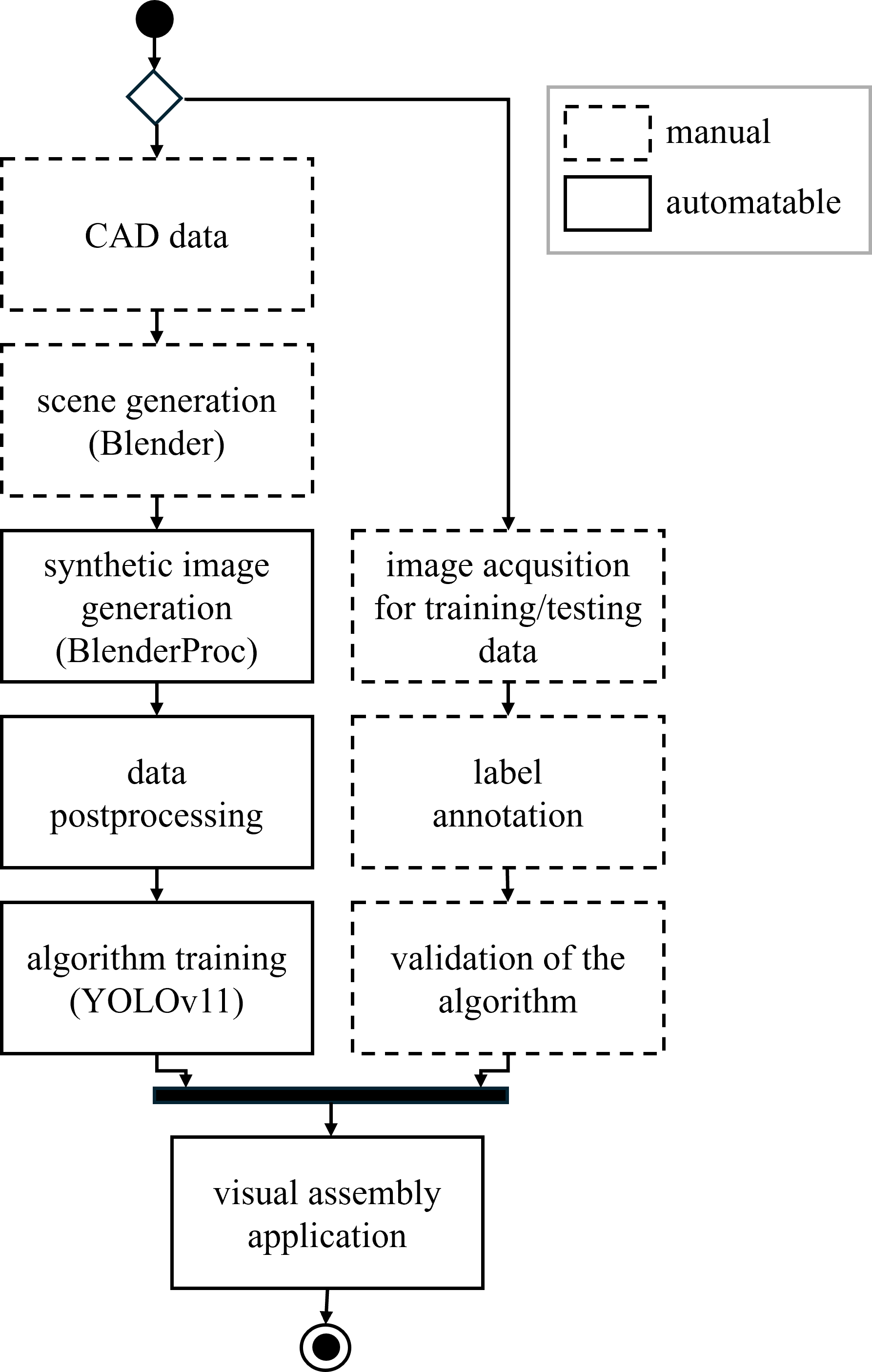}
        \caption{Synthetic data assembly pipeline modeled as UML activity diagram}
        \label{fig:pipeline}
\end{figure}

The visual assembly control pipeline consists of two parallel paths: a partially automated workflow for synthetic data generation and model training, and a manual workflow for real-world training and testing data evaluation. Figure \ref{fig:pipeline} illustrates the pipeline in Unified Modelling Language (UML) as an activity diagram \cite{Rumbaugh2004}. 

The process begins with CAD data of the assembly, which is often available in manufacturing companies. Alternatively, it can be manually created using CAD software. The data is then imported into Blender in formats such as Standard Triangle Language (STL) or STandard for the Exchange of Product data (STEP). Once imported, the CAD models serve as the basis for generating 3D scenes within Blender, forming the environment required for rendering synthetic image data.

Scene generation in Blender involves several key substeps including simulating a background plane that serves as the support surface for object placement, constructing scene collections that each comprise multiple components corresponding to annotated classes, and assigning material properties such as material type, color, and texture to all elements within the scene collections. Further details on these steps can be found in the official Blender documentation \cite{blender_manual}.

Next, BlenderProc is used to automatically generate synthetic images based on the created scene and predefined settings within a Python file, which specify parameters, such as lighting sources, object randomization within the scene, and the number of images to be generated. The background plane is defined as a passive rigid body, while other scene collections are assigned physical properties, enabling objects to interact under gravity and settle realistically onto the plane. Camera position determines the perspectives of generated images and light sources can be added with customizable numbers and intensities. Also with different conditions, such as as directional or diffuse light. The random positioning of objects can cause collisions, leading to unstable physics where components may be ejected out of the scene or become invisible in the rendered images behind other objects. The set number of images are generated with object annotations in COCO format for each object \cite{Lin14}. Besides, the defined number of generated images is accompanied by object annotations in COCO format for every object \cite{Lin14}.

The pipeline includes a postprocessing Python script that removes collided or invisible objects along with their automatically generated annotations, or deletes the entire image if no relevant objects remain. Furthermore, the postprocessing step includes converting annotations from COCO format to You Only Look Once (YOLO) annotation format for the subsequent processing stage. 

Afterwards, the synthetic data is processed for training a YOLO-series object detection model, with hyperparameters such as epochs and batch size configured within a Python file. YOLO series for object detection was chosen as the algorithmic foundation because it combines fast inference as a one-stage detector with ease of integration via its Application Programming Interface (API) \cite{redmon2016lookonceunifiedrealtime}. A pretrained YOLOv11m model, representing the current state of the art, is integrated into the training pipeline. The size of the YOLOv11m offers a good balance between accuracy and inference speed, enabling rapid model assembly. Moreover, the model can be easily switched within the YOLO series without requiring any code modifications. All steps from synthetic image generation to algorithm training are designed to be automatable, thereby minimizing manual effort and enhancing reproducibility. 

In parallel, real-world data need to be created manually for two purposes. Firstly, to develop a real-world model, and secondly, to compare this model with the synthetic model using additional unseen real-world testing images on both models. This includes image acquisition, manual label annotation, and algorithm validation with selected metrics. While these steps require human input, they are essential for assessing the model’s effectiveness under realistic conditions. Once the synthetic data has been validated, additional images can be generated without repeated testing.

Finally, the trained model is integrated to serve as a visual assembly control application, completing the pipeline.

\section{Experimental Use Case}
\label{setup}

In the following, the experimental use case is described along the steps of the pipeline. It involves two tasks for visual assembly control. First, individual parts of a selected assembly are detected. Second, another model determines whether these parts are assembled in a correct or incorrect configuration. Since both tasks are trained once on synthetic data and once on real-world data, we ultimately obtain four detectors: a synthetic-trained and a real-world-trained model for the assembly task, and a synthetic-trained and a real-world-trained model for the individual disassembled parts.

\begin{figure}
    \centering
    \begin{minipage}[b]{0.48\textwidth}
        \centering
        \includegraphics[width=\linewidth]{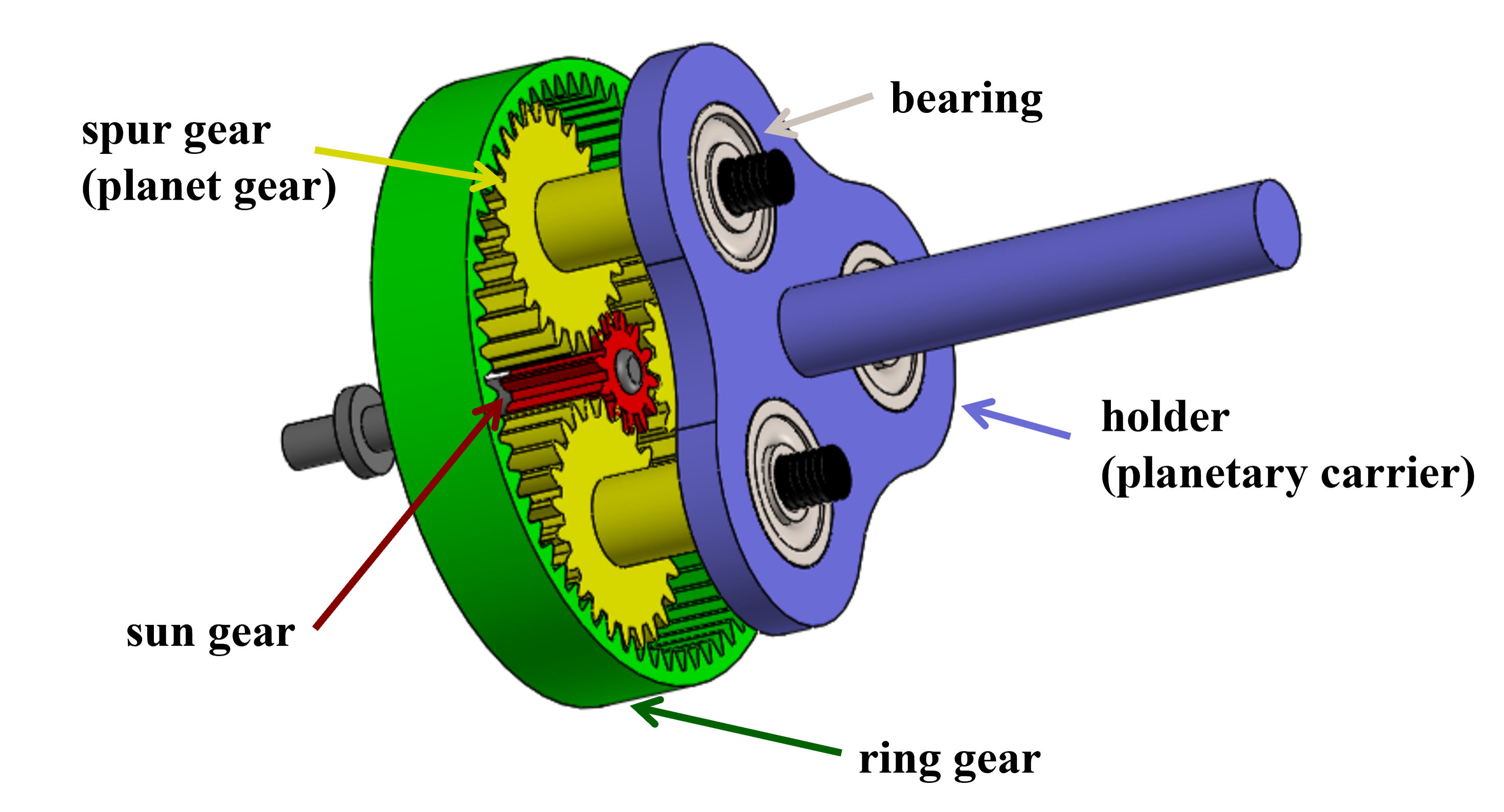}
        \caption{Planetary gear assembly in the CAD software}
        \label{fig:planetary_gear_system}
    \end{minipage}%
    \hfill
    \begin{minipage}[b]{0.48\textwidth}
        \centering
        \includegraphics[width=\linewidth]{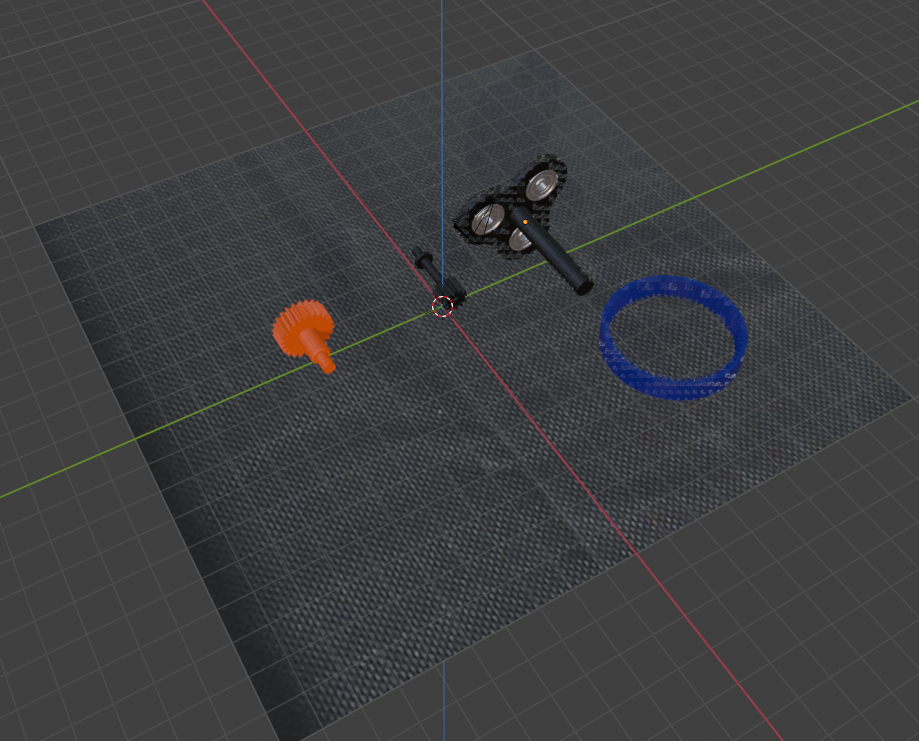}
        \caption{Planetary gear assembly components in Blender}
        \label{fig:Blender}
    \end{minipage}
\end{figure}

The selected exemplary object under inspection is a planetary gear assembly that consists of an additively manufactured holder, ring gear, sun gear, three spur gears, and three metal-based bearings that are not additively manufactured and installed in the holder. Thus, the assembly combines plastic and metal components. The geometrical file of this assembly is available open-source in the community GrabCAD \cite{Lis24} and illustrated in Figure \ref{fig:planetary_gear_system}.

The CAD data, specifically the planetary gear system’s STL files, are imported into Blender to construct the scene. The overall objective for the scene generation is to make all the components look as real as possible to ensure a high-fidelity representation of the real-world images. Therefore,  a background is modeled by generating a geometric plane whose surface is textured with an image reflecting the real-world environment where the objects are located. Afterwards, the objects are assigned to material properties, including color, and texture. For example, for the model of the additively manufactured plastic components, wave textures are selected to replicate the typical layer-by-layer pattern of 3D printing, while the bearings are assigned the material property of stainless steel. The color is assigned based on estimated RGB values. Figure \ref{fig:Blender} shows all parts in Blender with the real-world background image applied as a physical plane. The colors used for modeling the parts are based on the available 3D printer's filament colors and therefore differ from the coloration of the system shown in Figure \ref{fig:planetary_gear_system}.

\begin{figure}[htbp]
    \centering
    \begin{subfigure}[b]{0.45\textwidth}
        \centering
        \includegraphics[width=0.48\linewidth]{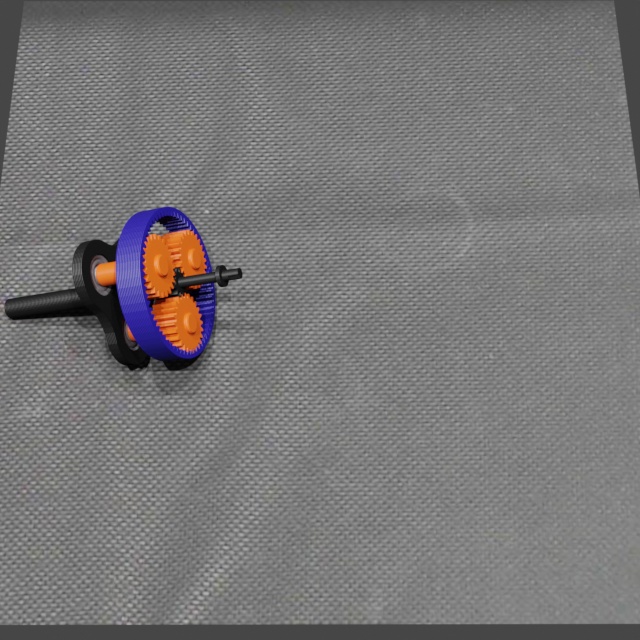}
        \includegraphics[width=0.48\linewidth]{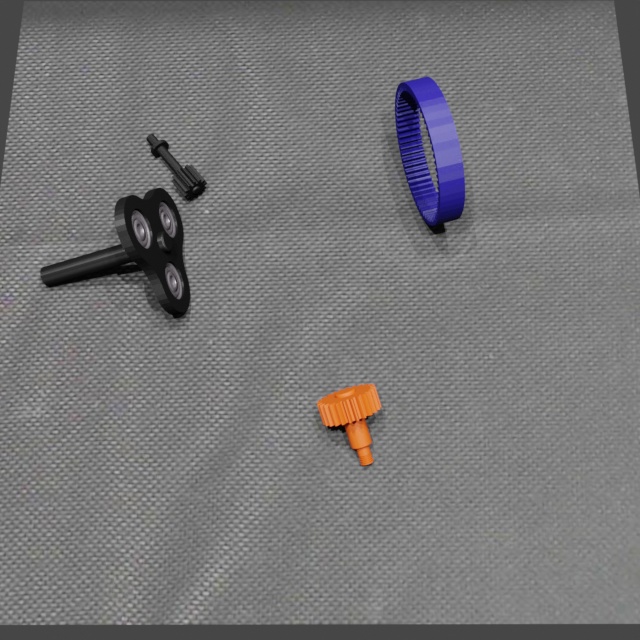}
        \caption{Synthetic images}
        \label{fig:synthetic_images}
    \end{subfigure}
    \hfill
    \begin{subfigure}[b]{0.45\textwidth}
        \centering
        \includegraphics[width=0.48\linewidth]{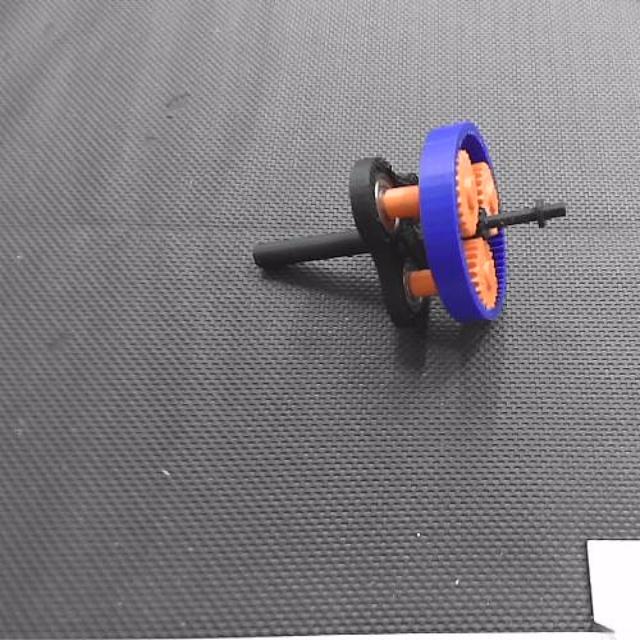}
        \includegraphics[width=0.48\linewidth]{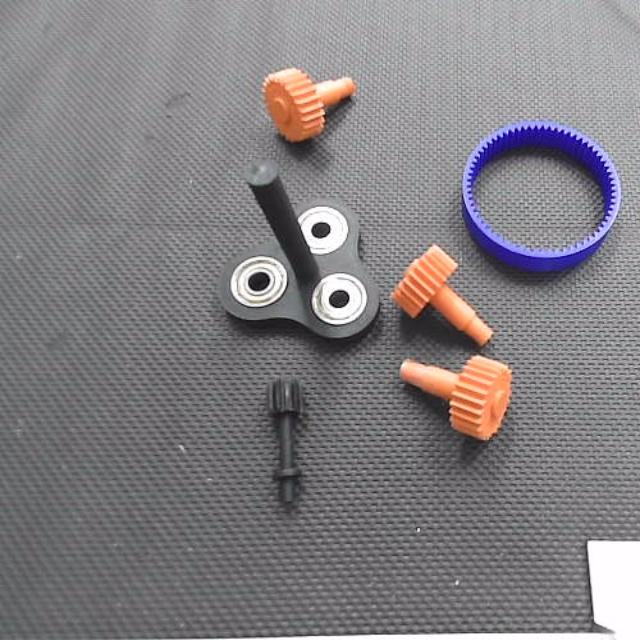}
        \caption{Real images}
        \label{fig:real_images}
    \end{subfigure}
    \caption{Comparison of synthetic images and real images for the disassembled (left) and assembled (right) planetary gear systems}
    \label{fig:synthetic_vs_real}
\end{figure}

Subsequently, images are generated using BlenderProc, based on specified camera parameters, lighting conditions, and the desired number of output images. The perspectives of objects in synthetic images are defined by the camera parameters. The camera position in the space is set with the format of three Cartesian position coordinates and three euler angles. The camera resolution is fixed at 640×640, as this resolution provides a good balance between computational efficiency and object detection accuracy for the YOLO-series algorithms \cite{7780460}. The position of light source is set in space with Cartesian coordinates as a directed light to replicate the directed lighting condition of the real scene. 
Each object in Blender has its category and is assigned to a certain class in annotation after generation. The plane serving the background needs to be set to a passive object, so that it is not assigned as a class for object detection. The active objects apply random positions and orientations in a customized range to ensure the diversity of generated images. Furthermore, the active objects are assigned to the physical characteristics influenced by properties such as gravity. The virtual scene and background plane are fixed in the Blender scene, while the activated objects (components of the planetary gear system) are positioned randomly within their physical space on the plane. The number of generated synthetic images is customized to 1043 for training and validation. The pipeline automatically renders and generates synthetic images with the annotation of all active objects. Figure \ref{fig:synthetic_vs_real} illustrates the synthetic images of the assembly and its parts along with real images of them in their actual environment.

In the context of assembly control, three classification categories are defined: "Qualified", "NoSunGear", and "NoHolder". Qualified means that all parts are correctly configured in the planetary gear system. To simulate a missing sun gear, the sun gear and ring gear are removed from one of the planetary gear systems. For the class indicating a missing holder, the holder is removed. Therefore, the scene contains three active objects corresponding to the three object detection classes, along with one passive object representing the background plane. For both models, the holder includes the bearing components.
In the case of the model for the disassembled individual components, four object categories are defined: "SunGear", "SpurGear", "Holder" (including bearings), and "GearRing". Moreover, in this scenario, each class is considered an active object, while the background plane remains the only passive object in the scene. A tutorial on scene and data creation is available in the supplementary materials of Werheid et al. \cite{WerheidGitLab2025}.

The 1043 generated synthetic images cover both use cases and are used for the training and validation datasets. Additionally, for each use case, 60 real-world images were manually captured and annotated using annotation software \cite{Dwyer2024}. Of these, 35 images were used to train and validate a comparable model, and the remaining 25 were used to test both the model trained on synthetic data and the model trained on real data. Consequently, two additional models, one for the assembled and one for the disassembled use-case, were trained and validated purely on the 35 real-world images.

The distribution of training, validation, and testing data, categorized by assembled and disassembled use cases, is summarized in Table \ref{tab:data_split}. YOLOv11m models are trained for each of the presented use cases. A 3.80 GHz CPU with 32 GB RAM was used, and the training times were 5h 48m for the synthetic-trained assembled model and 5h 39m for the synthetic-trained disassembled model.  In contrast, the real-world data models required significantly less training time due to its smaller dataset size.
A batch size of 32 and 100 epochs were selected based on previous tests and their loss curves. The AdamW optimizer was chosen based on recommendations from the library, which suggests using it for fewer than 10,000 iterations \cite{Jocher_Ultralytics_YOLO_2023}. 
The total number of iterations is calculated as follows: \[
\left.
\begin{array}{l}
\text{Total Iterations} = \left( \frac{\text{Training Dataset Size}}{\text{Batch size}} \right) \times \text{Epochs} \\
\phantom{\text{Total Iterations}} = \left( \frac{414}{32} \right) \times 100 = 1293.75
\end{array}
\right.
\]

The sizes of the other datasets are smaller, which results in a reduced number of iterations as well. The inference time for detection varies and is approximately 140 ms on the specified hardware for all trained models.

\begin{table}[ht]
  \caption{Dataset split (number of images) for each of the four YOLOv11 models}
  \label{tab:data_split}
  \centering
  \resizebox{\columnwidth}{!}{
  \begin{tabular}{|l|c|c|c|}
    \hline
    \textbf{Model} & \textbf{Training} & \textbf{Validation} & \textbf{Testing} \\
    \hline
    Assembled (synthetic)      & 400 (synthetic) & 101 (synthetic) & \multirow{4}{*}{25 (real)} \\
    Disassembled (synthetic)   & 414 (synthetic) & 128 (synthetic) & \\
    Assembled (real)           & 25 (real)       & 10 (real)       & \\
    Disassembled (real)        & 25 (real)       & 10 (real)       & \\
    \hline
  \end{tabular}}
\end{table}

The source code of the pipeline, along with an explanatory video demonstrating the scene generation process, is available online at \href{https://git.rwth-aachen.de/genai4zfp/synthetic_assembly_control}{https://git.rwth-aachen.de/genai4zfp/synthetic\_assembly\_control}~\cite{WerheidGitLab2025}. It is designed to be containerizable, enabling easy deployment across different computing environments. In support, a research data management plan based on RDMO~\cite{Windeck:918240} is provided in the repository to facilitate the FAIR (Findable, Accessible, Interoperable, Reusable) principles~\cite{Wilkinson2016}, enabling easier integration for industrial stakeholders to public data ~\cite{vlijmenetal20}.

\section{Results and Discussion}
\label{results}

To evaluate the model's performance, standard object detection metrics are used: Precision, recall, mAP@0.5, and mAP@0.5:0.95. Precision is defined as the ratio of true positives to the sum of true and false positives, while recall is the ratio of true positives to the sum of true positives and false negatives \cite{schlosser24}. The mean Average Precision (mAP) is calculated based on the Intersection over Union (IoU), which measures the overlap between the predicted and ground truth bounding boxes and is defined as the area of their intersection divided by the area of their union \cite{TALHA2023126881}. mAP@0.5 uses a fixed IoU threshold of 0.5, whereas mAP@0.5:0.95 averages the mAP over multiple IoU thresholds ranging from 0.5 to 0.95 in steps of 0.05. The model validation results for the synthetic trained models and the real-world trained models are summarized in Table \ref{val_synthetic} and Table \ref{val_real}, covering both the assembled and disassembled use cases.

\begin{table}
\caption{precision, recall, mAP@0.5 and mAP@0.5:0.95 of the synthetic-trained model on the synthetic validation dataset}
\centering
\resizebox{\columnwidth}{!}{
\begin{tabular}{|c|c|c|c|c|c|}
 \hline
                     & Class      & precision & recall   & mAP@0.5  & mAP@0.5:0.95\\ 
 \hline
                     & All        & 0.998     & 0.995    & 0.993    & 0.99\\
Assembled     & Qualified  & 0.998     & 0.986    & 0.988    & 0.988\\
                     & NoSunGear  & 0.999     & 1        & 0.995    & 0.987\\
                     & NoHolder   & 0.998     & 1        & 0.995   & 0.995\\
 \hline
                     & All        & 0.999     & 0.998    & 0.995    & 0.991\\
Disassembled  & SunGear    & 0.999     & 1        & 0.995    & 0.987\\
                     & SpurGear   & 1         & 0.991    & 0.995    & 0.989\\
                     & Holder     & 0.999     & 1        & 0.995    & 0.992\\
                     & GearRing   & 0.999     & 1        & 0.995    & 0.995\\
\hline 
\end{tabular}}
\label{val_synthetic}
\end{table}

\begin{table}
\caption{Precision, recall, mAP@0.5 and mAP@0.5:0.95 of the real-world trained model on the real-world validation dataset}
\centering
\resizebox{\columnwidth}{!}{
\begin{tabular}{|c|c|c|c|c|c|}
\hline
                     & Class        & Precision & Recall & mAP@0.5 & mAP@0.5:0.95 \\ 
\hline
                     & All          & 0.977     & 1.000  & 0.995   & 0.920 \\
Assembled            & Qualified    & 0.979     & 1.000  & 0.995   & 0.911 \\
                     & NoSunGear    & 0.961     & 1.000  & 0.995   & 0.927 \\
                     & NoHolder     & 0.992     & 1.000  & 0.995   & 0.921 \\
\hline
                     & All          & 0.992     & 1.000  & 0.995   & 0.923 \\
Disassembled         & SunGear      & 0.987     & 1.000  & 0.995   & 0.853 \\
                     & SpurGear     & 0.998     & 1.000  & 0.995   & 0.899 \\
                     & Holder       & 0.993     & 1.000  & 0.995   & 0.995 \\
                     & GearRing     & 0.990     & 1.000  & 0.995   & 0.946 \\
\hline 
\end{tabular}}
\label{val_real}
\end{table}

The validation results for precision, recall, mAP@0.5, and mAP@0.5:0.95 indicate good performance, suggesting that the synthetic models are suitable for making predictions on the synthetic validation datasets, and the real-world models on the real-world validation datasets. 

The results of these models on the real-world testing dataset are summarized in Table \ref{testing_synthetic} and Table \ref{test_real}.

\begin{table}
\caption{Precision, recall, mAP@0.5 and mAP@0.5:0.95 for the synthetic-trained model on real-world testing dataset}
\centering
\resizebox{\columnwidth}{!}{
\begin{tabular}{|c|c|c|c|c|c|}
 \hline
                     & Class      & Precision & Recall   & mAP@0.5  & mAP@0.5:0.95\\ 
 \hline
                     & All        & 0.925     & 0.812    & 0.926    & 0.784\\
Assembled     & Qualified  & 1         & 0.749    & 0.839    & 0.69\\
                     & NoSunGear  & 0.775     & 0.875    & 0.955    & 0.851\\
                     & NoHolder   & 1         & 0.812    & 0.984    & 0.812\\
 \hline
                     & All        & 0.964     & 0.987    & 0.991    & 0.765\\
Disassembled  & SunGear    & 0.982     & 1        & 0.995    & 0.536\\
                     & SpurGear   & 0.996     & 0.986    & 0.992    & 0.821\\
                     & Holder     & 0.9       & 0.96     & 0.981    & 0.773\\
                     & GearRing   & 0.977     & 1        & 0.995    & 0.93\\
\hline      
\end{tabular}}
\label{testing_synthetic}
\end{table}

\begin{table}
\caption{Precision, recall, mAP@0.5 and mAP@0.5:0.95 for the real-world trained model on real-world testing dataset}
\centering
\resizebox{\columnwidth}{!}{
\begin{tabular}{|c|c|c|c|c|c|}
\hline
                     & Class        & Precision & Recall & mAP@0.5 & mAP@0.5:0.95 \\ 
\hline
                     & All          & 0.868     & 0.806  & 0.864   & 0.768 \\
Assembled            & Qualified    & 0.684     & 0.750  & 0.845   & 0.675 \\
                     & NoSunGear    & 0.941     & 1.000  & 0.995   & 0.924 \\
                     & NoHolder     & 0.978     & 0.667  & 0.752   & 0.706 \\
\hline
                     & All          & 0.941     & 0.853  & 0.916   & 0.731 \\
Disassembled         & SunGear      & 0.940     & 0.624  & 0.839   & 0.499 \\
                     & SpurGear     & 0.990     & 0.986  & 0.995   & 0.844 \\
                     & Holder       & 0.993     & 0.800  & 0.838   & 0.695 \\
                     & GearRing     & 0.839     & 1.000  & 0.991   & 0.886 \\
\hline 
\end{tabular}}
\label{test_real}
\end{table}

\begin{figure}
  \centering
  \includegraphics[width=1\linewidth]{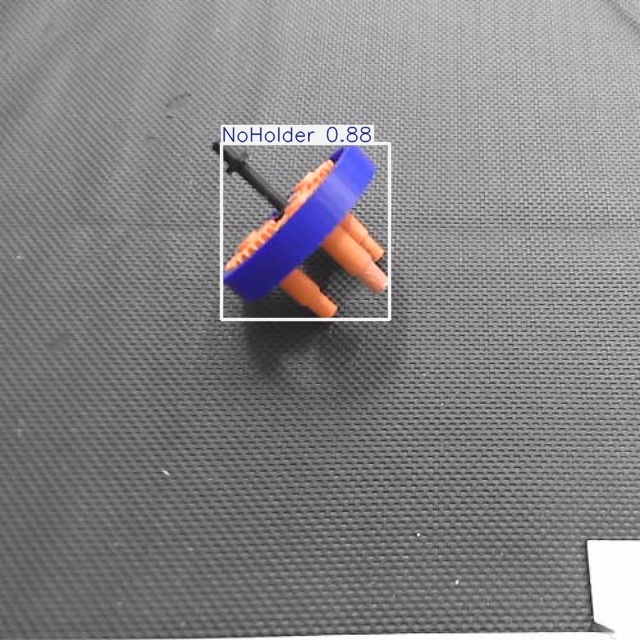}
  \caption{Detection of a misleading configuration (missing holder) in real-world data by the YOLOv11m  configuration model trained on synthetic data}
  \label{fig:example}
\end{figure}

The synthetic trained model of assembled components achieves an arithmetic mean \textit{mAP@0.5:0.95} of 0.784, while the disassembled model achieves an arithmetic mean \textit{mAP@0.5:0.95} of 0.765. Figure \ref{fig:example} illustrates an example of a misleading configuration in which the holder is missing, as detected by the synthetic-trained assembled model on real-world data.

Since the synthetic trained models record higher \textit{mAP@0.5:0.95} values on the validation dataset compared to the testing dataset, this indicates slight overfitting during training on the synthetic dataset. However, despite not having seen a single real image during training, the models generalize well to real-world testing data. 

The difference between the synthetic and real-world mAP@0.5:0.95 values is 0.206 for the synthetic-trained assembled model and 0.226 for the synthetic-trained disassembled model. These Sim2Real gaps are comparable to those reported in previous work \cite{Bai24}, which observed performance differences ranging from 0.017 to 0.318 using a YOLOv3 model for object detection. Additionally, the real-world models experience similar performance drops, with an average mAP@0.5:0.95 of 0.919 in the assembled validation dataset and 0.923 in the disassembled validation dataset, dropping to 0.768 and 0.729 in the testing dataset, respectively. It should be noted that the training data was relatively small, consisting of only 25 training instances and 10 validation instances. Besides, also the small size of the testing dataset limits the statistical significance of both findings as well. 

\begin{figure}[htbp]
    \centering
    \begin{subfigure}[b]{0.45\textwidth}
        \centering
        \includegraphics[width=0.48\linewidth]{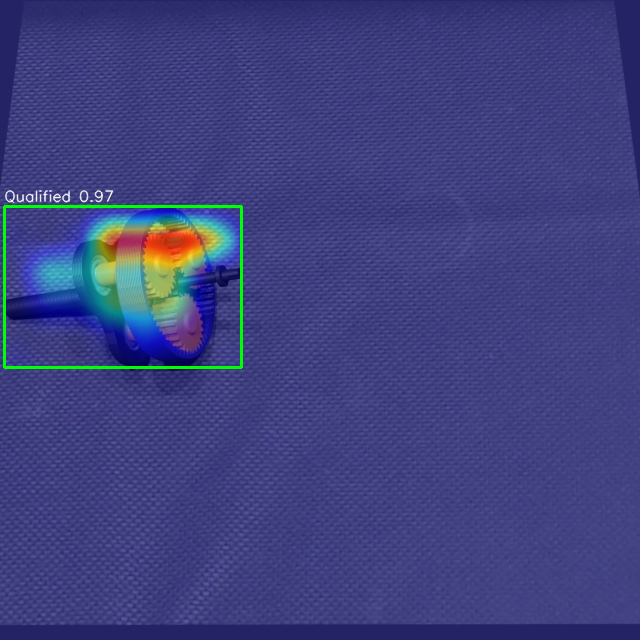}
        \includegraphics[width=0.48\linewidth]{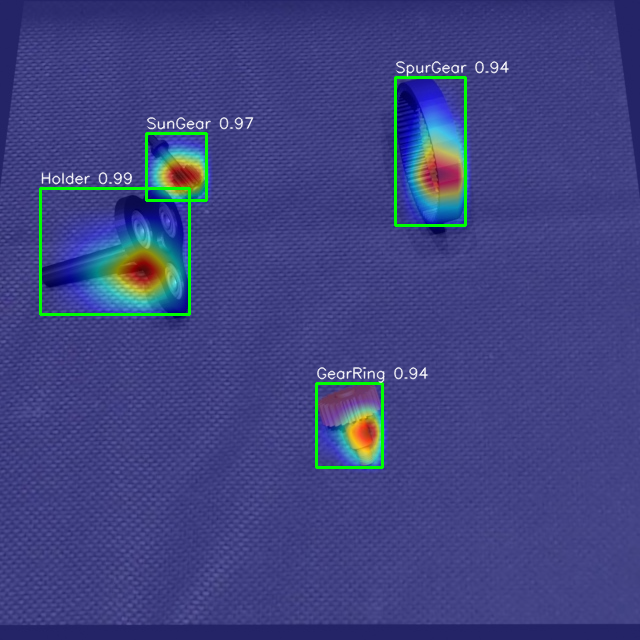}
        \caption{Synthetic images}
        \label{fig:synthetic_images}
    \end{subfigure}
    \hfill
    \begin{subfigure}[b]{0.45\textwidth}
        \centering
        \includegraphics[width=0.48\linewidth]{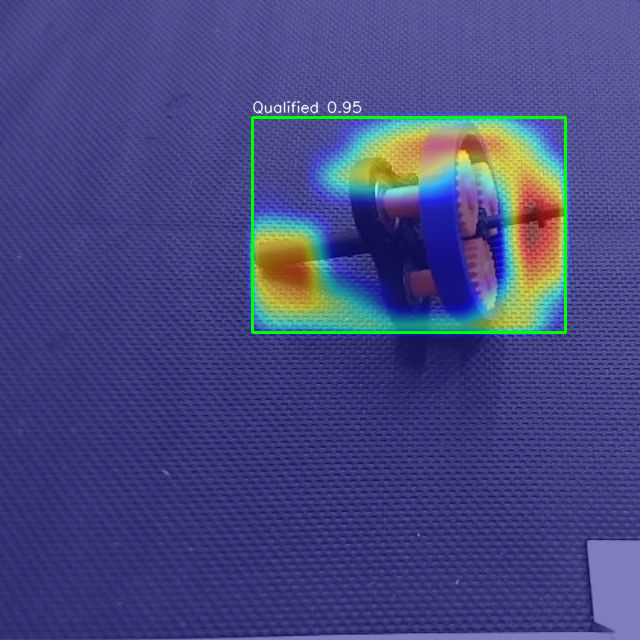}
        \includegraphics[width=0.48\linewidth]{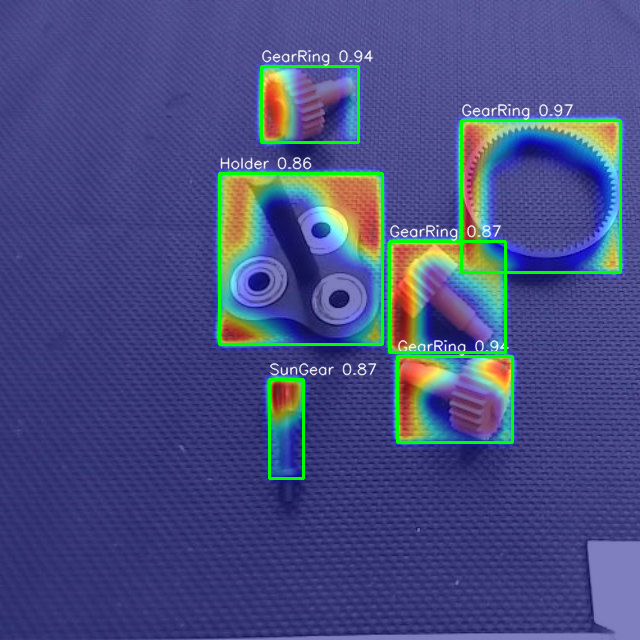}
        \caption{Real images}
        \label{fig:real_images}
    \end{subfigure}
    \caption{Comparison of predictions from the synthetic trained model on synthetic images (left) and real images (right) for the disassembled and assembled planetary gear systems, along with Eigen-CAM predictions based on \cite{rigved2024yolocam}}
    \label{fig:cam}
\end{figure}

To further interpret the predictions, Eigen Class Activation Mapping (Eigen-CAM) methods can be used to generate heatmaps that highlight important regions contributing to the image prediction. Figure \ref{fig:cam} illustrates heatmaps based on a Eigen-CAM algorithm tailored to YOLOv11 for the synthetic-trained models in assembled and disassembled predictions using both synthetic and real-world data \cite{rigved2024yolocam}.

It can be seen that in the analyzed examples, the predictions in real applications tend to be more influenced by the edges of the components, whereas in the synthetic example, they are more defined by the internal form elements. A systematic investigation into the causes of this effect is considered valuable for future initiatives.

\section{Conclusion}
\label{conclusion}

The control of assembly processes is a key factor in final product quality in manufacturing. Manual inspection by human operators remains time-consuming and involves repetitive tasks. While visual assembly control has been introduced, limited data availability is a common challenge, especially for smaller companies. 
This work presented a data-efficient and easily adaptable synthetic data pipeline for visual quality control in assembly processes. By leveraging CAD data, Blender, BlenderProc, and YOLOv11 object detection, we demonstrated that synthetic image datasets can be generated, annotated, and used to train models that achieved good accuracy on both synthetic validation and real-world data. Moreover, the integration into a single configurable pipeline, accompanied by source code and video tutorials, facilitates straightforward adoption by SMEs due to its simplicity. The results validated the feasibility of synthetic data for detecting correct and fault assembly configurations, with the synthetic trained models reaching a mAP@0.5:0.95 of up to 0.995 on the validation dataset and a mAP@0.5:0.95 up to 0.93 on real-world testing dataset, demonstrating similar performance to that of the real-world trained model. It should be noted that the statistical robustness is limited by the relatively small size of the real-world training and testing data. 
As the performance drop between validation and testing dataset indicated also a sim2real gap, future work could focus on improving realism and domain adaptation or investigating observed differences in prediction focus between real and synthetic data to further narrow the gap between synthetic training data and real-world testing images. This may include refining material properties, enhancing lighting simulation, or applying domain adaptation with generative models to bridge the Sim2Real gap using synthetic and unlabeled real data. Overall, the findings highlight the potential of simulation-based approaches to overcome common barriers to adopting industrial computer vision, such as limited data availability and costly manual annotation, by feasibly integrating synthetic data generation, model training, and deployment into a single, streamlined pipeline. This accessible and user-friendly solution is especially valuable for SMEs, empowering smaller-scale operations to innovate and remain competitive.



\section*{Acknowledgements}

The work is funded by the Federal Ministry for Economic Affairs and Energy - BMWE (GenAI4ZfP project, FKZ 01MK25005E). 



\section*{Contributions}

Jonas Werheid: Conceptualization, Data Analysis, Writing. Shengjie He: Data Analysis, Writing. Aymen Gannouni: Writing — Review. Anas Abdelrazeq: Writing — Review. Robert H. Schmitt: Supervision, Funding acquisition.

\bibliographystyle{IEEEtran}
\bibliography{sample}

\begin{thebibliography}{10}
\providecommand{\url}[1]{#1}
\csname url@samestyle\endcsname
\providecommand{\newblock}{\relax}
\providecommand{\bibinfo}[2]{#2}
\providecommand{\BIBentrySTDinterwordspacing}{\spaceskip=0pt\relax}
\providecommand{\BIBentryALTinterwordstretchfactor}{4}
\providecommand{\BIBentryALTinterwordspacing}{\spaceskip=\fontdimen2\font plus
\BIBentryALTinterwordstretchfactor\fontdimen3\font minus \fontdimen4\font\relax}
\providecommand{\BIBforeignlanguage}[2]{{%
\expandafter\ifx\csname l@#1\endcsname\relax
\typeout{** WARNING: IEEEtran.bst: No hyphenation pattern has been}%
\typeout{** loaded for the language `#1'. Using the pattern for}%
\typeout{** the default language instead.}%
\else
\language=\csname l@#1\endcsname
\fi
#2}}
\providecommand{\BIBdecl}{\relax}
\BIBdecl

\bibitem{nof1997industrial}
\BIBentryALTinterwordspacing
S.~Y. Nof, W.~E. Wilhelm, and H.-J. Warnecke, \emph{Industrial Assembly}, 1st~ed.\hskip 1em plus 0.5em minus 0.4em\relax New York, NY: Springer, 1997. [Online]. Available: \url{https://doi.org/10.1007/978-1-4615-6393-8}
\BIBentrySTDinterwordspacing

\bibitem{Beiter00}
K.~Beiter, B.~Cheldelin, and K.~Ishii, ``Assembly quality method: A tool in aid of product strategy, design, and process improvements,'' 09 2000, pp. 149--156.

\bibitem{BBD22}
F.~P. Basamakis, A.~C. Bavelos, D.~Dimosthenopoulos, A.~Papavasileiou, and S.~Makris, ``Deep object detection framework for automated quality inspection in assembly operations,'' \emph{Procedia CIRP}, vol. 115, pp. 166--171, 2022.

\bibitem{ZAMORAHERNANDEZ2021103485}
\BIBentryALTinterwordspacing
M.-A. Zamora-Hernández, J.~A. Castro-Vargas, J.~Azorin-Lopez, and J.~Garcia-Rodriguez, ``Deep learning-based visual control assistant for assembly in industry 4.0,'' \emph{Computers in Industry}, vol. 131, p. 103485, 2021. [Online]. Available: \url{https://www.sciencedirect.com/science/article/pii/S0166361521000920}
\BIBentrySTDinterwordspacing

\bibitem{Basamakis22}
F.~Basamakis, A.~Bavelos, D.~Dimosthenopoulos, A.~Papavasileiou, and S.~Makris, ``Deep object detection framework for automated quality inspection in assembly operations,'' \emph{Procedia CIRP}, vol. 115, pp. 166--171, 11 2022.

\bibitem{eng2.12910}
\BIBentryALTinterwordspacing
J.~Werheid, S.~Münker, N.~Klasen, T.~Hamann, A.~Abdelrazeq, and R.~H. Schmitt, ``Demonstrating computer vision to small- and medium-sized enterprises in manufacturing: Toward overcoming costs and implementation challenges,'' \emph{Engineering Reports}, vol.~6, no.~11, p. e12910, 2024. [Online]. Available: \url{https://onlinelibrary.wiley.com/doi/abs/10.1002/eng2.12910}
\BIBentrySTDinterwordspacing

\bibitem{Werheid2025}
\BIBentryALTinterwordspacing
J.~Werheid, H.~Behnen, J.-H. Woltersmann, S.~He, T.~Hamann, A.~Abdelrazeq, and R.~H. Schmitt, ``Machine vision in manufacturing smes: a review,'' \emph{Discover Applied Sciences}, vol.~7, no.~5, p. 371, 2025. [Online]. Available: \url{https://doi.org/10.1007/s42452-025-06923-4}
\BIBentrySTDinterwordspacing

\bibitem{Buggineni22}
V.~Buggineni, C.~Chen, and J.~Camelio, ``Enhancing manufacturing operations with synthetic data: A systematic framework for data generation, accuracy, and utility,'' \emph{Frontiers in Manufacturing Technology}, 01 2024.

\bibitem{Qian22}
C.~Qian, W.~Yu, C.~Lu, D.~Griffith, and N.~Golmie, ``Toward generative adversarial networks for the industrial internet of things,'' \emph{IEEE Internet of Things Journal}, vol.~9, pp. 1--1, 10 2022.

\bibitem{Ameperosa20}
E.~Ameperosa and P.~Bhounsule, ``Domain randomization using deep neural networks for estimating positions of bolts,'' \emph{Journal of Computing and Information Science in Engineering}, vol.~20, pp. 1--18, 05 2020.

\bibitem{LAI202069}
\BIBentryALTinterwordspacing
Z.-H. Lai, W.~Tao, M.~C. Leu, and Z.~Yin, ``Smart augmented reality instructional system for mechanical assembly towards worker-centered intelligent manufacturing,'' \emph{Journal of Manufacturing Systems}, vol.~55, pp. 69--81, 2020. [Online]. Available: \url{https://www.sciencedirect.com/science/article/pii/S0278612520300303}
\BIBentrySTDinterwordspacing

\bibitem{blender18}
\BIBentryALTinterwordspacing
B.~O. Community, \emph{Blender - a 3D modelling and rendering package}, Blender Foundation, Stichting Blender Foundation, Amsterdam, 2018. [Online]. Available: \url{http://www.blender.org}
\BIBentrySTDinterwordspacing

\bibitem{MONNET2024767}
\BIBentryALTinterwordspacing
J.~Monnet, O.~Petrovic, and W.~Herfs, ``Investigating the generation of synthetic data for surface defect detection: A comparative analysis,'' \emph{Procedia CIRP}, vol. 130, pp. 767--773, 2024, 57th CIRP Conference on Manufacturing Systems 2024 (CMS 2024). [Online]. Available: \url{https://www.sciencedirect.com/science/article/pii/S2212827124013192}
\BIBentrySTDinterwordspacing

\bibitem{SONG2024109852}
\BIBentryALTinterwordspacing
J.~Song, X.~Qin, J.~Lei, J.~Zhang, Y.~Wang, and Y.~Zeng, ``A fault detection method for transmission line components based on synthetic dataset and improved yolov5,'' \emph{International Journal of Electrical Power \& Energy Systems}, vol. 157, p. 109852, 2024. [Online]. Available: \url{https://www.sciencedirect.com/science/article/pii/S0142061524000735}
\BIBentrySTDinterwordspacing

\bibitem{Károly22}
A.~I. Károly and P.~Galambos, ``Automated dataset generation with blender for deep learning-based object segmentation,'' in \emph{2022 IEEE 20th Jubilee World Symposium on Applied Machine Intelligence and Informatics (SAMI)}, 2022, pp. 000\,329--000\,334.

\bibitem{LV2024102788}
\BIBentryALTinterwordspacing
N.~Lv, D.~Zhao, F.~Kong, Z.~Xu, and F.~Du, ``A multi-feature fusion-based pose tracking method for industrial object with visual ambiguities,'' \emph{Advanced Engineering Informatics}, vol.~62, p. 102788, 2024. [Online]. Available: \url{https://www.sciencedirect.com/science/article/pii/S1474034624004361}
\BIBentrySTDinterwordspacing

\bibitem{Den2023}
\BIBentryALTinterwordspacing
M.~Denninger, D.~Winkelbauer, M.~Sundermeyer, W.~Boerdijk, M.~Knauer, K.~H. Strobl, M.~Humt, and R.~Triebel, ``Blenderproc2: A procedural pipeline for photorealistic rendering,'' \emph{Journal of Open Source Software}, vol.~8, no.~82, p. 4901, 2023. [Online]. Available: \url{https://doi.org/10.21105/joss.04901}
\BIBentrySTDinterwordspacing

\bibitem{CONRAD2024239}
\BIBentryALTinterwordspacing
J.~Conrad, T.~Stauffer, X.~Meng, J.~Ferchow, and M.~Meboldt, ``Deep learning-based error recognition in manual cable assembly using synthetic training data,'' \emph{Procedia CIRP}, vol. 128, pp. 239--244, 2024, 34th CIRP Design Conference. [Online]. Available: \url{https://www.sciencedirect.com/science/article/pii/S2212827124006668}
\BIBentrySTDinterwordspacing

\bibitem{Biruduganti25}
S.~Biruduganti, Y.~Yardi, and L.~Ankile, ``Bridging the sim2real gap: Vision encoder pre-training for visuomotor policy transfer,'' 01 2025.

\bibitem{Bai24}
K.~Bai, L.~Zhang, Z.~Chen, F.~Wan, and J.~Zhang, ``Close the sim2real gap via physically-based structured light synthetic data simulation,'' 07 2024.

\bibitem{RAWAL20251668}
\BIBentryALTinterwordspacing
P.~Rawal, M.~Sompura, and W.~Hintze, ``Synthetic data generation procedures for domain-specific environments in manufacturing,'' \emph{Procedia Computer Science}, vol. 253, pp. 1668--1679, 2025, 6th International Conference on Industry 4.0 and Smart Manufacturing. [Online]. Available: \url{https://www.sciencedirect.com/science/article/pii/S1877050925002376}
\BIBentrySTDinterwordspacing

\bibitem{WU2022138}
\BIBentryALTinterwordspacing
C.~Wu, K.~Zhou, J.-P. Kaiser, N.~Mitschke, J.-F. Klein, J.~Pfrommer, J.~Beyerer, G.~Lanza, M.~Heizmann, and K.~Furmans, ``Motorfactory: A blender add-on for large dataset generation of small electric motors,'' \emph{Procedia CIRP}, vol. 106, pp. 138--143, 2022, 9th CIRP Conference on Assembly Technology and Systems. [Online]. Available: \url{https://www.sciencedirect.com/science/article/pii/S221282712200169X}
\BIBentrySTDinterwordspacing

\bibitem{BURGGRAEF2024775}
\BIBentryALTinterwordspacing
P.~Burggraef, T.~Adlon, F.~Steinberg, F.~Broehl, A.~Moriz, C.~Engeln, M.~Schütz, and F.~Jaworek, ``Automatic generation of assembly instructions by analyzing process recordings – a concept overview,'' \emph{Procedia CIRP}, vol. 126, pp. 775--780, 2024, 17th CIRP Conference on Intelligent Computation in Manufacturing Engineering (CIRP ICME ‘23). [Online]. Available: \url{https://www.sciencedirect.com/science/article/pii/S2212827124009144}
\BIBentrySTDinterwordspacing

\bibitem{Rumbaugh2004}
J.~Rumbaugh, I.~Jacobson, and G.~Booch, \emph{Unified Modeling Language Reference Manual, The (2nd Edition)}.\hskip 1em plus 0.5em minus 0.4em\relax Pearson Higher Education, 2004.

\bibitem{blender_manual}
\BIBentryALTinterwordspacing
{Blender Foundation}, ``Blender manual,'' 2024, accessed: 2025-06-08. [Online]. Available: \url{https://docs.blender.org/manual/en/latest/}
\BIBentrySTDinterwordspacing

\bibitem{Lin14}
T.-Y. Lin, M.~Maire, S.~Belongie, J.~Hays, P.~Perona, D.~Ramanan, P.~Dollár, and C.~Zitnick, ``Microsoft coco: Common objects in context,'' vol. 8693, 04 2014.

\bibitem{redmon2016lookonceunifiedrealtime}
\BIBentryALTinterwordspacing
J.~Redmon, S.~Divvala, R.~Girshick, and A.~Farhadi, ``You only look once: Unified, real-time object detection,'' 2016. [Online]. Available: \url{https://arxiv.org/abs/1506.02640}
\BIBentrySTDinterwordspacing

\bibitem{Lis24}
\BIBentryALTinterwordspacing
{Lishan Isuru}, ``Planetary gear system | 3d cad model library | grabcad,'' 2024. [Online]. Available: \url{https://grabcad.com/library/planetary-gear-system-16}
\BIBentrySTDinterwordspacing

\bibitem{7780460}
J.~Redmon, S.~Divvala, R.~Girshick, and A.~Farhadi, ``You only look once: Unified, real-time object detection,'' in \emph{2016 IEEE Conference on Computer Vision and Pattern Recognition (CVPR)}, 2016, pp. 779--788.

\bibitem{WerheidGitLab2025}
J.~Werheid, ``Synthetic assembly control — software, data, and explanation videos,'' \url{https://git.rwth-aachen.de/genai4zfp/synthetic_assembly_control}, 2025, accessed: 2025-06-02.

\bibitem{Dwyer2024}
\BIBentryALTinterwordspacing
B.~Dwyer, J.~Nelson, T.~Hansen \emph{et~al.}, ``Roboflow,'' 2024, computer vision software. [Online]. Available: \url{https://roboflow.com}
\BIBentrySTDinterwordspacing

\bibitem{Jocher_Ultralytics_YOLO_2023}
\BIBentryALTinterwordspacing
G.~Jocher, J.~Qiu, and A.~Chaurasia, ``{Ultralytics YOLO},'' Jan. 2023. [Online]. Available: \url{https://github.com/ultralytics/ultralytics}
\BIBentrySTDinterwordspacing

\bibitem{Windeck:918240}
\BIBentryALTinterwordspacing
J.~Windeck, D.~Wallace, D.~A. Hausen, and S.~Schönau. (2022) {RDMO} : {E}in {W}erkzeug zur {P}lanung und {U}msetzung des {F}orschungsdatenmanagements. Zuerst gesehen am 25.09.2023. [Online]. Available: \url{https://publications.rwth-aachen.de/record/918240}
\BIBentrySTDinterwordspacing

\bibitem{Wilkinson2016}
\BIBentryALTinterwordspacing
M.~D. Wilkinson, M.~Dumontier, I.~J. Aalbersberg, G.~Appleton, M.~Axton, A.~Baak, N.~Blomberg, J.-W. Boiten, L.~B. da~Silva~Santos, P.~E. Bourne, J.~Bouwman, A.~J. Brookes, T.~Clark, M.~Crosas, I.~Dillo, O.~Dumon, S.~Edmunds, C.~T. Evelo, R.~Finkers, A.~Gonzalez-Beltran, A.~J. Gray, P.~Groth, C.~Goble, J.~S. Grethe, J.~Heringa, P.~A. ’t Hoen, R.~Hooft, T.~Kuhn, R.~Kok, J.~Kok, S.~J. Lusher, M.~E. Martone, A.~Mons, A.~L. Packer, B.~Persson, P.~Rocca-Serra, M.~Roos, R.~van Schaik, S.-A. Sansone, E.~Schultes, T.~Sengstag, T.~Slater, G.~Strawn, M.~A. Swertz, M.~Thompson, J.~van~der Lei, E.~van Mulligen, J.~Velterop, A.~Waagmeester, P.~Wittenburg, K.~Wolstencroft, J.~Zhao, and B.~Mons, ``The fair guiding principles for scientific data management and stewardship,'' \emph{Scientific Data}, vol.~3, no.~1, p. 160018, Mar. 2016. [Online]. Available: \url{https://doi.org/10.1038/sdata.2016.18}
\BIBentrySTDinterwordspacing

\bibitem{vlijmenetal20}
H.~van Vlijmen, A.~Mons, A.~Waalkens, W.~Franke, A.~Baak, G.~Ruiter, C.~Kirkpatrick, L.~O.~B. da~Silva~Santos, B.~Meerman, R.~Jellema, D.~Arts, M.~Kersloot, S.~Knijnenburg, S.~Lusher, R.~Verbeeck, and J.-M. Neefs, ``Your article title here,'' \emph{Journal Name}, vol.~X, no.~Y, pp. ZZZ--ZZZ, 2020.

\bibitem{schlosser24}
T.~Schlosser, M.~Friedrich, T.~Meyer, and D.~Kowerko, ``A consolidated overview of evaluation and performance metrics for machine learning and computer vision,'' 01 2024.

\bibitem{TALHA2023126881}
\BIBentryALTinterwordspacing
M.~M. Talha, H.~U. Khan, S.~Iqbal, M.~Alghobiri, T.~Iqbal, and M.~Fayyaz, ``Deep learning in news recommender systems: A comprehensive survey, challenges and future trends,'' \emph{Neurocomputing}, vol. 562, p. 126881, 2023. [Online]. Available: \url{https://www.sciencedirect.com/science/article/pii/S0925231223010044}
\BIBentrySTDinterwordspacing

\bibitem{rigved2024yolocam}
R.~Sharan, ``Yolo-v12-cam: Eigencam for yolo v11 interpretability,'' \url{https://github.com/rigvedrs/YOLO-V12-CAM}, 2024, accessed: 2025-07-23.

\end{thebibliography}
\end{document}